\documentclass[MSc]{icldt}
\usepackage{amsthm}
\usepackage{hyperref}
\hypersetup{
    colorlinks,
    citecolor=black,
    filecolor=black,
    linkcolor=black,
    urlcolor=black
}
\usepackage[british]{babel}
\usepackage{hyphenat}
\usepackage{graphicx}
\usepackage{amsmath}
\usepackage{algorithmicx}
\usepackage{algpseudocode}
\usepackage[chapter]{algorithm}
\usepackage{svg}

\title{Comparison of Training Methods for Deep Neural Networks}
\author{Patrick Oliver GLAUNER}

\date{April 2015}
\department{Computing}
\field{Computing (Machine Learning)}
\supervisor{Professor Maja PANTIC \\ and Dr. Stavros PETRIDIS}
\declaration{I herewith certify that all material in this report which is not my own work has been properly acknowledged.}

\begin{document}

\maketitle

\begin{abstract}
This report describes the difficulties of training neural networks and in particular deep neural networks. It then provides a literature review of training methods for deep neural networks, with a focus on pre-training. It focuses on Deep Belief Networks composed of Restricted Boltzmann Machines and Stacked Autoencoders and provides an outreach on further and alternative approaches. It also includes related practical recommendations from the literature on training them.
In the second part, initial experiments using some of the covered methods are performed on two databases. In particular, experiments are performed on the MNIST hand-written digit dataset and on facial emotion data from a Kaggle competition. The results are discussed in the context of results reported in other research papers.
An error rate lower than the best contribution to the Kaggle competition is achieved using an optimized Stacked Autoencoder.

\end{abstract}


\tableofcontents
\listoftables
\listoffigures

\chapter{Introduction}
\label{chapter:intro}
Neural networks have a long history in machine learning. Early experiments have shown both, their expressional power, but also the difficulty to train them.
For the last ten years, neural networks are celebrating a comeback under the label "deep learning".
Enormous efforts in research have been made on this topic, which attracted major IT companies including Google, Facebook, Microsoft and Baidu to make significant investments in deep learning.

Deep learning is not simply a revival of an old theory, but it comes with completely different ways of building and training many-layer neural networks. This rise has been supported by recent advances in computer engineering, in particular the strong parallelism in CPUs and GPUs.

Most prominently, the so-called "Google Brain project" has been in the news for its capability to self-learn cat faces from images extracted from YouTube videos as presented in \cite{google_brain}.
Aside from computer vision, major advances in machine learning have been reported in audio and natural language processing.
These advances have been raising many hopes about the future of machine learning, in particular to work   towards building a system that implements the single learning hypothesis as presented by Ng in \cite{ng_deep}.

Nonetheless, deep learning has not been reported to be a easy and quick to use silver bullet to any machine learning problem.
In order to apply the theoretical foundations of deep learning to concrete problems, much experimentation is required.

Given the success reported in many applications, deep learning looks promising to be applied to various computer vision problems, such as hand-written digit recognition and facial expression recognition. For the latter, only few results have been reported so far, such as in \cite{liu1} and \cite{liu2}, and to intelligent behavior understanding as whole. An example is the work in \cite{tang}, which won the 2013 Kaggle facial expression competition \cite{kaggle_competition} using a deep convolutional neural network.

\chapter{Neural networks}
\label{chapter:nn}
This chapter provides a summary of neural networks, their history and training challenges. Prior exposure to neural networks is assumed, this chapter is therefore not considered to give an introduction to neural networks.

The perceptron introduced by Rosenblatt in 1962 \cite{rosenblatt} is a linear, non\hyp{}differentiable classifier, making it of very limited use on real data. Quite related to perceptrons, logistic regression is a linear classifier, whose cost function is differentiable.
Inspired by the brain, neural networks are composed of layers of perceptrons or logistic regression units.

\section{Feed-forward neural networks}
\label{chapter:nn:feed}
\begin{figure}[h!]
    \centering
    \includegraphics[width=0.6\textwidth]{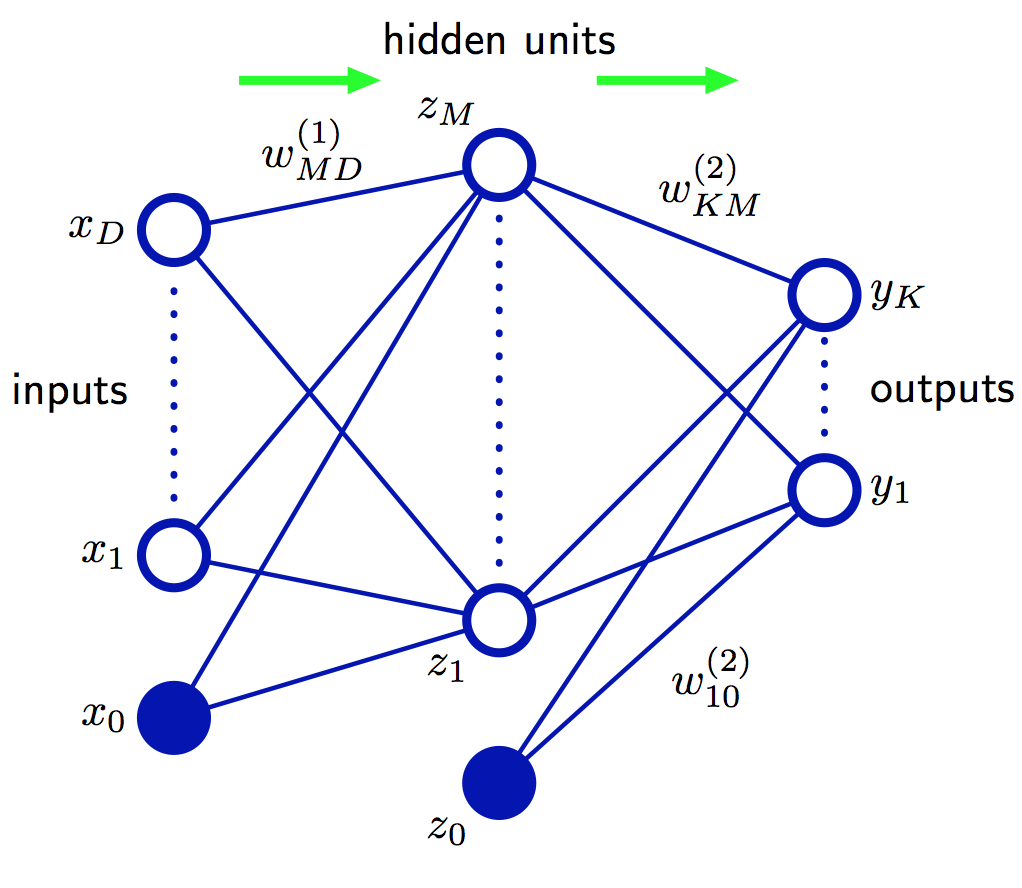}
    \caption{Neural network with two input and output units and one hidden layer with two units and bias units $x_0$ and $z_0$ \cite{bishop}}
    \label{fig:NN}
\end{figure}

In their simplest form, feed-forward neural networks propagate inputs through the network to make a prediction, whose output is either continuous or discrete, for regression or classification, respectively. A simple neural network is visualized in Figure~\ref{fig:NN}.

Using learned weights $\Theta$, the activation of unit $i$ of layer $j+1$ can be calculated:
\begin{align}
a_i^{(j+1)} = g\left(\sum_{k=0}^{s_{j}} \Theta_{ik}^{(j)}x_k\right)
\end{align}
$g$ is an activation function. Most commonly, the Sigmoid activation function $\frac{1}{1 + e^{-x}}$ is used for classification problems.

Given the limitations of a single perceptron unit, \cite{perceptrons} led to a long-standing debate of perceptrons and to a misinterpretation of the power of neural networks. As a consequence, research interest and funding in research on neural networks dropped for about two decades.

Another downside of neural networks is learning the weights between the large amounts of parameters and the required computational resources, which is covered in Chapter~\ref{chapter:nntraining}.
In the 1980s, regained attention in research led to backpropagation, an efficient training algorithm for neural networks.

Subsequent research led to very successful applications, such as the following examples.
The Mixed National Institute of Standards and Technology (MNIST) database \cite{MNIST} is a large collection of handwritten digits with different levels of noise and distortions.
Classifying handwritten digits is needed in different tasks, for example in modern and highly automated mail delivery. Neural networks were found in the 1980s to perform very well on this task.
Autonomous driving is subject to current research in AI. Back in the 1980s, significant steps in this field were made using neural networks as shown by Pomerleau in \cite{ALVINN}.

Neural networks are known to learn complex non-linear hypotheses very well, leading to low training errors. Nonetheless, training a neural network is non-trivial. Given properly extracted features as input, most classification problems can be done with one hidden layer quite accurately as described in \cite{heaton}. Adding another layer allows to learn even more complex hypotheses for scattered classes.
There are many different (and contradictory) approaches to finding appropriate network architectures as summarized and discussed in \cite{stackoverflow_nn}.
In general, the more units and layers, the higher a network's expressional power. This makes training more difficult.

Extracting proper features is a major difficulty in signal processing and computer vision as presented by Ng in \cite{ng_deep}.
Neural networks with many hidden layers allow to learn features from a dataset themselves, coming with enormous simplifications in those fields.
The computational expensiveness and the low generalization capabilities of deep neural networks resulted in another decline of neural networks in the 1990s as described in \cite{deng}.

Many-layer neural networks, so-called deep neural networks, have been receiving lots of attention for about the last ten years because of new training methods, which are covered in Chapter~\ref{chapter:deepnn}. These developments have achieved significant progress in machine learning, as described in Chapter~\ref{chapter:intro} and in \cite{hinton_science}, \cite{hinton_speech} and \cite{speech_spectrogram}.
In addition to new training methods, advances in computer engineering, in particular faster CPUs and GPUs, have supported the rise of deep neural networks. GPUs have proved to significantly speed up training of deep neural networks, as reported in \cite{gpu_coates}.
This has also been raising many hopes about the future of machine learning to push its frontiers towards building a system that implements the single learning hypothesis as presented by Ng in \cite{ng_deep}.

\begin{figure}[h!]
    \centering
    \includegraphics[width=0.8\textwidth]{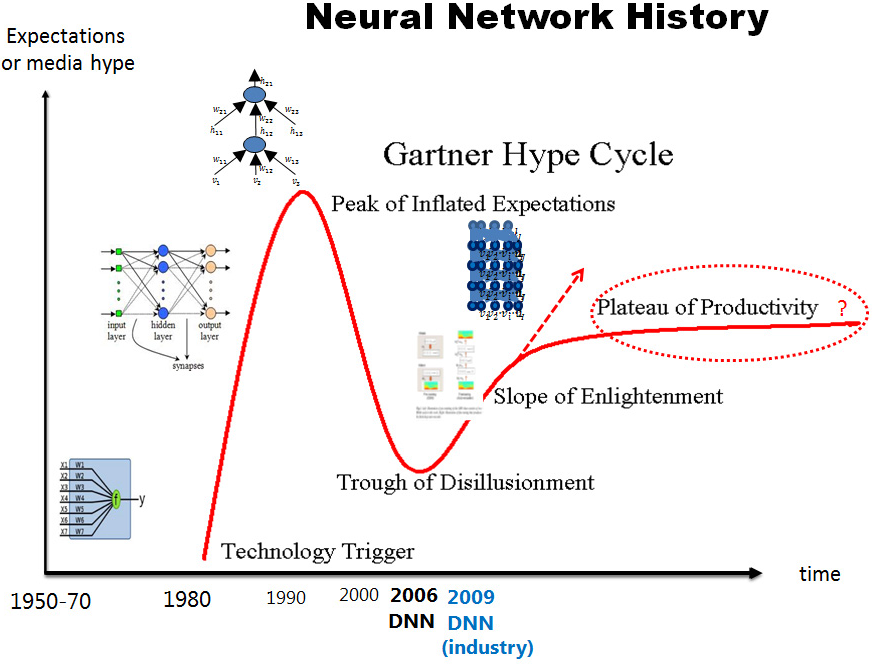}
    \caption{History of neural networks \cite{deng}}
    \label{fig:nnhype}
\end{figure}

As visualized in Figure~\ref{fig:nnhype}, the history of neural networks can be plotted on a hype curve with open questions and resulting uncertainty of the future of neural networks.

\section{Other types of neural networks}
\label{chapter:nn::other}
In addition to feed-forward networks, there are two other popular types of neural networks: convolutional and recurrent neural networks \cite{bishop}. Convolutional neural networks (CNN) are a special kind of feed-forward networks that include invariance properties in their architecture, for example to tolerate translation of the input.
Their structure implies constraints on the weights, as weights are shared within so-called feature maps. This approach reduces the number of independent parameters and therefore speeds up learning using minor modifications in backpropagation.
CNNs have been used successfully in computer vision, for example on MNIST.
Recurrent neural networks (RNNs) are cyclic graphs of units. This architecture results in an internal state of the network, allowing it to exhibit dynamic temporal behavior. In practice, training RNNs can come with complex graph problems, making them difficult to understand and train reliably.

Stochastic neural networks covered in \cite{nn_stochastic} are special type of neural networks. They include random variations in either their weights or in their activation function. They allow to escape from local optima. Boltzmann Machines and Restricted Boltzmann Machines presented in Chapter~\ref{chapter:deepnn} are stochastic neural networks.

\section{Training neural networks}
\label{chapter:nntraining}
In order to train a neural network, its prediction error is measured using a cost function, for which there are many different ones in the literature, such as least squares cost function
\begin{align}
J(\Theta) = \sum_{i = 1}^m (y^{(i)} - h_\Theta(x^{(i)}))^2
\end{align}
and cross-entropy cost function covered in \cite{mitchell}, which is meant to generalize better and speed up learning as discussed in \cite{ng_mlc}:
\begin{align}
J(\Theta) = -\left (\sum_{i = 1}^m y^{(i)} \log h_\Theta(x^{(i)}) + (1 - y^{(i)}) \log (1-h_\Theta(x^{(i)})) \right)
\end{align}

Cost functions of neural networks are highly non-convex as visualized by Ng in \cite{ng_mlc}.
Therefore, finding the global minimum of the cost function is difficult and optimization often converges in a local minimum.

In order to minimize the error function, the partial derivatives of the network parameters are calculated:
The simplest method to approximate the gradient of any cost function is to utilize finite differences:
\begin{align}
\frac{\partial}{\partial\theta_i}J(\theta)\approx\frac{J(...,\theta_{i-1},\theta_{i}+\epsilon,\theta_{i+1},...)-J(\theta)}{\epsilon}
\end{align}

Alternatively, symmetric differences allow higher precision:
\begin{align}
\frac{\partial}{\partial\theta_i}J(\theta)\approx\frac{J(...,\theta_{i-1},\theta_{i}+\epsilon,\theta_{i+1},...)-J(...,\theta_{i-1},\theta_{i}-\epsilon,\theta_{i+1},...)}{2\epsilon}
\end{align}

Both methods are computationally inefficient, as the cost function needs to be evaluated twice per partial derivative.
In the 1980s, backpropagation was developed, which is an efficient method to compute the partial derivatives \cite{mitchell}.
Its main idea is to start with random initial weights between the units and then to propagate training data through the network. Initial weights are random, in order to avoid symmetrical behavior of the neurons that would lead to an inefficient and redundant network behavior. The magnitude of error is then used to layer-wise backpropagate derivatives of the error, which are finally used to update the weights. This leads to a reduction of the training error.

Next, the partial derivatives are used in an optimization algorithm to minimize the cost function, such as in gradient descent defined in Algorithm~\ref{alg:gradientdescent}.
Each update of gradient descent goes through the entire training set of $m$ training examples, making single updates expensive. Furthermore, gradient descent does not allow to escape local minima, as each update steps goes further down the cost function towards a minimum.

\begin{algorithm}
\caption{Batch gradient descent: training size $m$, learning rate $\alpha$}
\label{alg:gradientdescent}
\begin{algorithmic}
\Repeat
\State $\theta_j  \gets \theta_j - \alpha \frac{\partial}{\partial \theta_j}J(\theta)$ (simultaneously for all $j$)
\Until{convergence}
\end{algorithmic}
\end{algorithm}

In practice, preferably stochastic gradient descent, also named online gradient descent, defined in Algorithm~\ref{alg:stochasticgradientdescent} is used in neural networks, as explained by LeCun in \cite{lecun_research}. It wanders around minima and lacks certain convergence criteria of gradient descent. For example, some update steps might go away from the nearest minimum or even up on the cost function. In contrast, each update step is for a single training example, making updates much faster. Also, stochastic gradient descent allows to escape local minima under certain circumstances.

\begin{algorithm}
\caption{Stochastic gradient descent: training size $m$, learning rate $\alpha$}
\label{alg:stochasticgradientdescent}
\begin{algorithmic}
\State Randomly shuffle data set
\Repeat
\For{$i=1$ to $m$}
\State $\theta_j  \gets \theta_j - \alpha \frac{\partial}{\partial \theta_j}J(\theta, (x^{(i)}, y^{(i)}))$ (simultaneously for all $j$)
\EndFor
\Until{convergence}
\end{algorithmic}
\end{algorithm}

\subsection{Difficulty of training neural networks}

In general, the more units and layers, the higher a network's expressional power. This comes with a more complex  cost function. Learning will then easily get stuck in a local minimum, leading to high generalization errors and overfitting. There are different methods to reduce overfitting, for example smaller networks, more training data, better optimization algorithms, or regularization, which is covered in Chapter~\ref{chapter:nnreg}.

Deep learning utilizes deep neural networks consisting of many layers of neurons. In principle, backpropgation works on such networks as well. Nonetheless, it does not perform very well on large networks because of the following considerations:
For small initial weights, the partial derivative updates back propagated through many layers will be very small or 0 because of numerical effects, leading to very unsatisfiable updates. This is also known as the vanishing gradient problem.
On the other hand, large initial weights will make the training get stuck from the beginning on in some region of the cost function, not allowing to converge to the global or better local minimum, which is covered in detail in \cite{difficulty_training} and \cite{pretraining_help}.

\section{Regularization}
\label{chapter:nnreg}
Regularization allows to reduce overfitting of neural networks and learning algorithms in general.
This section covers different regularization methods.

\subsection{$L_2$ and $L_1$ regularization}
Large parameter values often allow models to match training data well, but do not generalize to new data.
$L_2$ and $L_1$ regularization "penalize" large parameter values, by adding them to the cost function.

For the weights, the following notation of Ng in \cite{ng_mls} is used: $\Theta^{(l)}$ is a matrix of weights to map from layer $l$ to layer $l+1$.
If the network has $s_l$ units in layer $l$ and $s_{l+1}$ units in layer $l+1$, then $\Theta^{(l)}$ is of dimension $s_{l+1}\times (s_l+ 1)$, also considering the extra bias unit. 

$L_2$ regularization also known as weight decay is used predominantly and adds the squared weights to the cost function:
\begin{align}
J_{reg}(\Theta) = J(\Theta)+ \lambda \sum_{l=1}^{L-1}\sum_{i=1}^{s_{l+1}}\sum_{j=1}^{s_{l}}\left(\Theta_{ij}^{(l)}\right)^2
\end{align}
The magnitude of the regularization can be controlled by the regularization parameter $\lambda$, whose value is often subject to model selection.

An alternative approach is $L_1$ regularization, which only adds the linear weights to the cost function:
\begin{align}
J_{reg}(\Theta) = J(\Theta)+ \lambda \sum_{l=1}^{L-1}\sum_{i=1}^{s_{l+1}}\sum_{j=1}^{s_{l}} \left\lvert \Theta_{ij}^{(l)}\right\rvert
\end{align}

\begin{figure}[h!]
    \centering
    \includegraphics[width=0.8\textwidth]{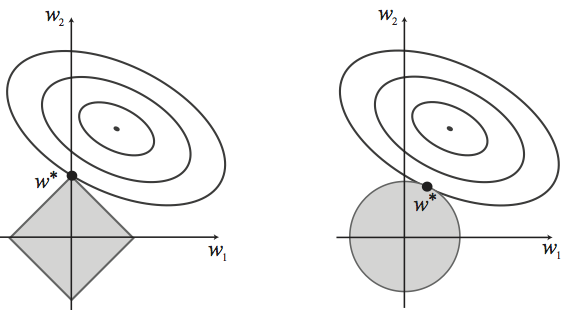}
    \caption{$L_1$ and $L_2$ regularization \cite{norvig}}
    \label{fig:l1l2}
\end{figure}

Both $L_1$ and $L_2$ regularization have been extensively compared by Ng in \cite{ng_l1_l2}.
$L_1$ regularization tends to produce a sparse model, which often sets many parameters to zero, effectively declaring the corresponding attributes to be irrelevant. $L_2$ regularization gives no preference to zero weights.
This behavior is visualized in Figure~\ref{fig:l1l2}.

\subsection{Early stopping}
Another popular regularization method is early stopping, in which for every training iteration, not only the training error, but also a validation error is calculated. The training error is usually a monotonic function, that decreases further in every iteration.
In contrast, the validation error usually drops off first, but then increases, indicating the model starting to overfit.
In early stopping, training is stopped at the lowest error on the validation set.

\subsection{Invariance}
Neural networks can be made invariant of transformations, such as rotation or translation. A simple method is to apply these transformations on the training data and then to train the neural network on the modified training data as well.
A different approach is tangent propagation \cite{bishop} which embeds a regularization function into the cost function. The regularization function is high for non-invariant network mapping functions and zero when the network is invariant under the transformation.

\subsection{Dropout}
\label{chapter:deepnn:dropout}
\begin{figure}[h!]
    \centering
    \includegraphics[width=0.9\textwidth]{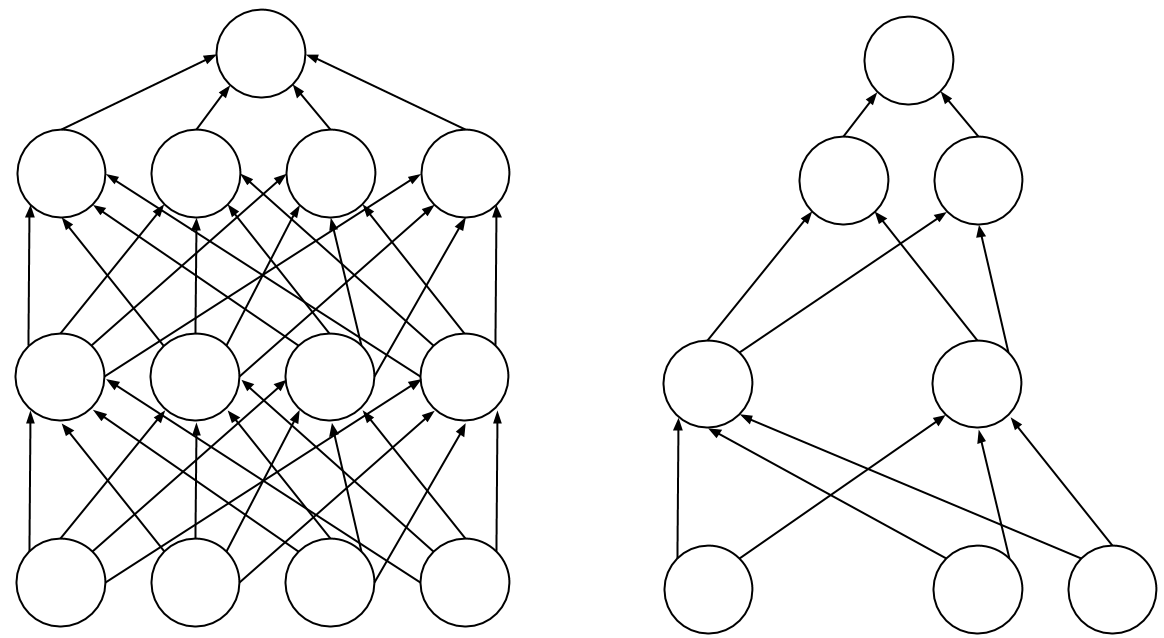}
    \caption{Left: a neural network with two hidden layers. Right: a result of applying dropout to the same neural network.}
    \label{fig:dropout}
\end{figure}

Dropout regularization presented in \cite{dropout_simple} is a general form of regularization for neural networks. It is motivated by genetic algorithms covered in \cite{genetic_mitchell}, which combine the genes of parents and apply some mutation in order to produce an offspring.
The idea behind dropout regularization is to thin a neural network by dropping out some of its units, as visualized in Figure~\ref{fig:dropout}.

The lower number of parameters reduces overfitting of the network.
Each unit is retained with probability $p$, where $p$ is fixed and set for all units. $p$ can be found in model selection. Alternatively, it can be set to 0.5, which is a near-optimal choice for most applications, as evaluated in \cite{dropout_simple}. It should be set to greater than 0.5 and closer to 1 for the input units, in order to preserve most input units. It has been empirically determined that a value of 0.8 is a good value for the input units.

For a network with $n$ units, there are $2^n$ different thinned networks, which share most weights, and the total number of parameters is still $O(n^2)$.
For each training example, all thinned networks are sampled and trained. 
The weights of the thinned networks can be combined in order to construct a network with $n$ units. This process allows to filter out noise in the training data, in order to prevent overfitting.

This approach can be interpreted as adding noise to the network's hidden units, a form of regularization. It is related to the implicit regularization in denoising autoencoders and tangent propagation, presented in Chapters~\ref{chapter:autoenc:den} and \ref{chapter:nnreg}, respectively.
In contrast to the deterministic noise in denoising autoencoders, dropout is a stochastic regularization.

\chapter{Deep neural networks}
\label{chapter:deepnn}
Given the limitations of training deep neural networks described in Chapter~\ref{chapter:nn}, this chapter presents various methods and algorithms that allow to overcome these limitations in order to construct powerful deep neural networks.

Designing features has been a difficult topic in signal processing. For example, the SIFT detector defined by Lowe in \cite{lowe} is a powerful feature sector, but it is difficult to adapt to specific problems in computer vision.
A different approach is to learn features from data. This can be achieved by designing feature detectors in a way to model the structure of the input data well.
The first step of deep learning is generative pre-training, which works in a way to subsequently learn layers of feature detectors. Starting with simple features, these serve as input to the next layer in order to learn more complex features.

Pre-training allows to find a good initialization of the weights, which is a region of the cost function that can be optimized quickly through discriminative fine-tuning of the network using backpropoagation.

\section{Restricted Boltzmann machines}
\label{chapter:rbm}
A Boltzmann Machine is a generative recurrent stochastic neural network that allows to learn complex features from data. As intra-layer connections make learning difficult and inefficient, it is not discussed further.

\begin{figure}[h!]
    \centering
    \includegraphics[width=0.9\textwidth]{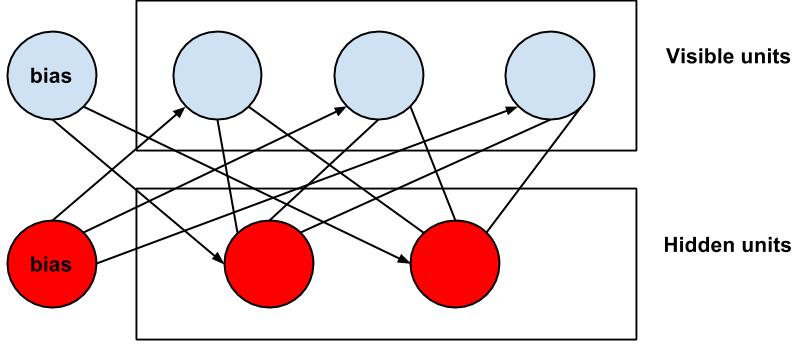}
    \caption{Restricted Boltzmann Machine with three visible units and two hidden units (and biases)}
    \label{fig:RBM}
\end{figure}

A Restricted Boltzmann Machine (RBM) defined by Hinton in \cite{hinton_guide} and \cite{hinton_science} is a Boltzmann Machine, in which which the neurons are binary nodes of a bipartite graph. (A bipartite graph is a graph whose nodes can be grouped into two disjoint sets, such that every edge connect a node from the first set to the other.)
Both groups of neurons/units are called visible and hidden layer as visualized in Figure~\ref{fig:RBM}.
The general concept of RBMs is demonstrated in \cite{edwinchen}.

The visible units of a RBM represent states that are observed, i.e. input to the network, whereas the hidden units represent the feature detectors. RBMs are undirected, with a single matrix $\mathbf{W}$ of parameters, which associates the connectivity of visible units $\mathbf{v}$ and hidden units $\mathbf{h}$. Furthermore, there are bias units $\mathbf{a}$ for the visible units and $\mathbf{h}$ for the hidden units.

The energy of a joint configuration ($\mathbf{v}$, $\mathbf{h}$) of the units of both layers has an energy function:
\begin{align}
E(\mathbf{v}, \mathbf{h}) = -\sum_i a_i v_i - \sum_j b_j h_j -\sum_i \sum_j v_i w_{ij} h_j
\end{align}
expressed in matrix notation:
\begin{align}
E(\mathbf{v}, \mathbf{h}) = -a^Tv - b^Th - v^TWh
\end{align}

This energy function is derived in \cite{stochastic_methods} for simulated annealing, an optimization algorithm. Its inspiration comes from annealing in metallurgy.
The optimization goal is to change the weights so that desirable configurations have low energy.

Probability distributions over joint configurations are defined in terms of the energy function:
\begin{align}
p(\mathbf{v}, \mathbf{h}) = \frac{1}{Z}e^{-E(\mathbf{v}, \mathbf{h})}
\end{align}
$Z$ is a normalizer, also called partition function, which is the sum over all possible configurations:
\begin{align}
Z = \sum_{\mathbf{v}, \mathbf{h}}e^{-E(\mathbf{v}, \mathbf{h})}
\end{align}

Subsequently, the probability assigned by the network to a visible vector is the marginal probability:
\begin{align}
p(\mathbf{v}) = \frac{1}{Z}\sum_{\mathbf{h}}e^{-E(\mathbf{v}, \mathbf{h})}
\end{align}

Since the RBM is a bipartite graph and has therefore no intra-layer connections, a binary hidden unit is set to 1 with the following probability for a input vector $\mathbf{v}$:
\begin{align}
\label{eq:hidden}
p(h_j = 1 \vert \mathbf{v}) = \sigma(b_j + \sum_i v_i w_{ij})
\end{align}

$\sigma(x)$ is the Sigmoid activation function $\frac{1}{1 + e^{-x}}$.
In the following notation, $<v_i h_j>_{distribution}$ denotes an expectation under a distribution.
The expectation observed in the training set $<v_i h_j>_{data}$ is then easy to get, where $h_j$ sampled given $v_i$.
Given the bipartite graph, the same considerations apply to the probability of a visible unit set to 1 for a hidden vector $\mathbf{h}$:

\begin{align}
\label{eq:visible}
p(v_i = 1 \vert \mathbf{h}) = \sigma(a_i + \sum_j h_j w_{ij})
\end{align}
In contrast, getting an unbiased sample under the the distribution defined by the model $<v_i h_j>_{model}$ is computationally inefficient as explained by Hinton in \cite{hinton_guide}, \cite{hinton_shapes} and \cite{hinton_science}.

\subsection{Training}
\label{chapter:deepnn:naive}
Learning the connectivity weights is performed in a way to learn features detectors in the hidden layer of the visible layer.
More precisely, the optimization objective of RBMs is to maximize the weights and biases in order to assign high probabilities to training examples and to lower the energy of these training examples.
Conversely, any other examples shall have a high energy.

The derivative of the log probability of a training example with respect to a weight $w_{ij}$ between input $i$ and feature detector $j$ is as shown by Hinton in \cite{hinton_guide}:
\begin{align}
\frac{\partial \log p(\mathbf{v})}{\partial w_{ij}} = <v_i h_j>_{data} - <v_i h_j>_{model}
\end{align}

A simple rule for performing stochastic gradient descent is then with learning rate $\alpha$:
\begin{align}
\Delta w_{ij} = \alpha (<v_i h_j>_{data} - <v_i h_j>_{model})
\end{align}

As mentioned previously, sampling $<v_i h_j>_{model}$ is computationally inefficient, making this training algorithm infeasible.

\subsection{Contrastive divergence}
\label{chapter:deepnn:cd}
Contrastive divergence covered by Hinton in \cite{hinton_guide} approximates the gradient doing the following procedure, assuming that all visible and hidden units are binary:
First, the visible units are set to a training example. Second, the binary states of the hidden layer are computed using \ref{eq:hidden}. Third, a binary so-called "reconstruction" of the visible layer is computed using \ref{eq:visible}.
Last, \ref{eq:hidden} is applied again to compute binary values of the hidden layer.
The weight update rule is then:
\begin{align}
\Delta w_{ij} = \alpha (<v_i h_j>_{data} - <v_i h_j>_{recon})
\end{align}

The reason for using binary and not probabilistic values in the hidden units comes from information theory. Feature detectors serve as information bottlenecks, as each hidden units can transmit at least one bit. Using probabilities in the hidden units would violate this bottleneck.
For the visible units, also probabilistic values could be used with little consequences.

In general, contrastive divergence is much faster than the basic training algorithm presented previously and returns usually well-trained RBM weights.

\section{Autoencoders}
\begin{figure}[h!]
    \centering
    \includegraphics[width=0.9\textwidth]{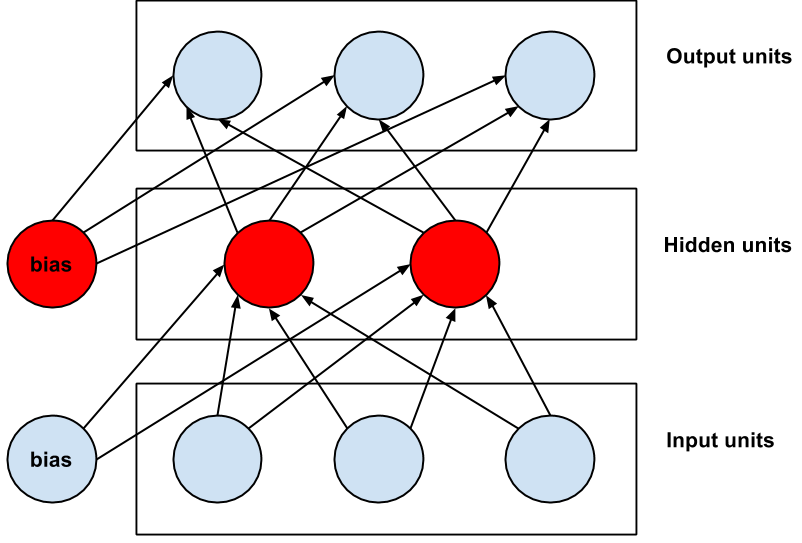}
    \caption{Autoencoder with three input and output units and two hidden units}
    \label{fig:autencoder}
\end{figure}

An autoencoder or autoassociator described in \cite{ng_tutorial} and \cite{deng} is a three-layer neural network with $s_1$ input and output units and $s_2$ hidden units as visualized in Figure~\ref{fig:autencoder}.

It sets its output values to its input values, i.e. $y^{(i)} = x^{(i)}$ and tries to learn the identity function $h_\Theta(x) \approx x$. This can be achieved using backpropagation in its basic definition.
After learning the weights, the hidden layer units become feature detectors.
The number of hidden units is variable. If $s_2 < s_1$, the autoencoder performs dimensionality reduction, as shown in \cite{learning_deep} and \cite{speech_spectrogram}.
In contrast, $s_2 > s_1$ maps the input to a higher dimension.

\subsection{Sparse autoencoders}
\label{chapter:autoenc::sparse}
For a large number of hidden units, a sparse autoencoder defined in \cite{ng_tutorial} still discovers interesting features in the input by putting a sparsity constraint on the weights. This constraint only allows a small number of units to be activated for every input vector.
Sparsity in autoencoders may be constructed using $L_1$ regularization as described in Chapter~\ref{chapter:nnreg}.
An alternative method uses the Kullback-Leibler (KL) divergence, which is a measure of how different two distributions are, given by a Bernoulli random variable with mean $p$ and a Bernoulli random variable with mean $q$:
\begin{align}
KL(p\Vert q) = p \log \frac{p}{q} + (1-p) \log \frac{1-p}{1-q} 
\end{align}
The KL divergence is 0 for $p = q$, otherwise positive.

In the following definition, the activation of hidden unit $j$ for input $x^{(i)}$ is written $a^{(2)}_j(x^{(i)})$.
The average activation of unit $j$ for a training set is defined as follows:
\begin{align}
\hat p_j = \frac{1}{m} \sum_{i=1}^m a^{(2)}_j(x^{(i)})
\end{align}

In order to compute $\hat p$, the entire training set needs to be propagated forward first in the autoencoder.
The sparsity optimization objective for unit $j$ is:
\begin{align}
\hat p_j = p
\end{align}
where $p$ is the sparsity parameter, which is usually a small value. For example, for $p = 0.05$, the average activation of unit $j$ is 5\%.
In order to enforce an average activation $p$ for all hidden units, the following regularization term can be added to the cost function:
\begin{align}
J_{reg}(\Theta) = J(\Theta) + \lambda \sum_{j}^{s_2}KL(p\Vert \hat p_j)
\end{align}
The regularization part adds a positive value to the cost function for $p \neq \hat p_j$ and the global minimum for sufficient $\lambda$ and therefore satisfies a sparse model.

This method is statistically more accurate than $L_1$ regularization, but requires more complex changes to backpropagation as shown in \cite{ng_tutorial}. It is computationally more expensive because of the extra forward propagation of the entire training set.

\subsection{Denoising autoencoders}
\label{chapter:autoenc:den}
An denoising autoencoder defined in \cite{stacked_denoising} is an autoencoder that was trained to denoise corrupted inputs.
In the training process, outputs $y^{(i)}$ are obtained by corrupting the corresponding inputs using a deterministic corruption mapping: $y^{(i)} = f_\Theta(x^{(i)})$.
Using backpropagation, the autoencoder learns denoising, i.e. mapping a corrupted example back to an uncorrupted one. From a geometric perspective, this is also called manifold learning and related to tangent propagation mentioned in Chapter~\ref{chapter:nnreg}.

There are multiple proposals on how to modify the cost function to learn different degrees of denoising.
One of the more complex denoising methods presented in \cite{stacked_denoising} also handles salt-and-pepper-noise. The idea behind it is to penalize the projection errors of each dimension differently, proportionally to the degree of noise in each dimension.

\section{Comparison of RBMs and autoencoders}
\label{chapter:rbmautoenc}
RBMs and autoencoders are two different approaches in order to build Deep Belief Networks introduced in Chapter~\ref{chapter:deepbelief}.
It is generally difficult to determine which of them is better, as both have been applied to different use cases under different situations in the literature.
Nonetheless, the mathematical foundations of autoencoders are simpler than RBMs.
This chapter does not attempt to determine the better approach, but rather describes the differences of their optimization problems based on \cite{rbmvsauto_charles}:

RBMs and autoencoders both solve an optimization problem of the form:
\begin{align}
\min J_{recon}(\theta) + regularization
\end{align}
i.e. optimizing the reconstruction error $J_{recon}(\theta)$ and some form of regularization.

In the basic training of RBMs described in Chapter~\ref{chapter:deepnn:naive}, the optimization objective of RBMs is to minimize the reconstruction error.
This naive training would require to fully evaluate $\log Z$, where $Z$ is the partioning function.
Since this is impractical, contrastive divergence described in Chapter~\ref{chapter:deepnn:cd} only partially evaluates $Z$, which is the form of regularization of this method.

In contrast, the optimization problem of autoencoders is to reduce the reconstruction error with a sparse model:
\begin{align}
\min J_{recon}(\theta) + sparsity
\end{align}
For the sparsity, there are a variety of regularization terms such as $L_1$ or KL divergence as covered in Chapter~\ref{chapter:autoenc::sparse}.

\section{Deep belief networks}
\label{chapter:deepbelief}
A Deep Belief Network (DBN) is a stack of simple networks, such as RBMs or autoencoders, that were trained layer-wise in an unsupervised procedure.
In the following section, RBMs are used to explain the idea behind DBNs. DBNs composed of autoencoders are also called stacked autoencoders, which are explained in Chapter~\ref{chapter:deepnn:stacked}.

\begin{figure}[h!]
    \centering
    \includegraphics[width=0.3\textwidth]{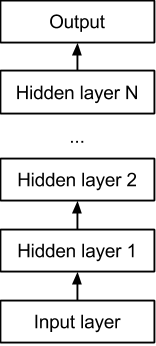}
    \caption{Deep belief network structure}
    \label{fig:DBN_schema}
\end{figure}

Training of a DBN consists of two stages, that allows to learn feature hierarchies, as described in \cite{hinton_science}, \cite{stacked_denoising} and \cite{exploring_strategies}.
In the first stage, generative unsupervised learning is performed layer-wise on RBMs.
First, a RBM is trained on the data. Second, its hidden units are used as input to another RBM, which is trained on them. This process can be continued for multiple RBMs, as visualized in Figure~\ref{fig:DBN_schema}.
As a consequence, each RBM learns more complex features.

\begin{figure}[h!]
    \centering
    \includegraphics[width=0.9\textwidth]{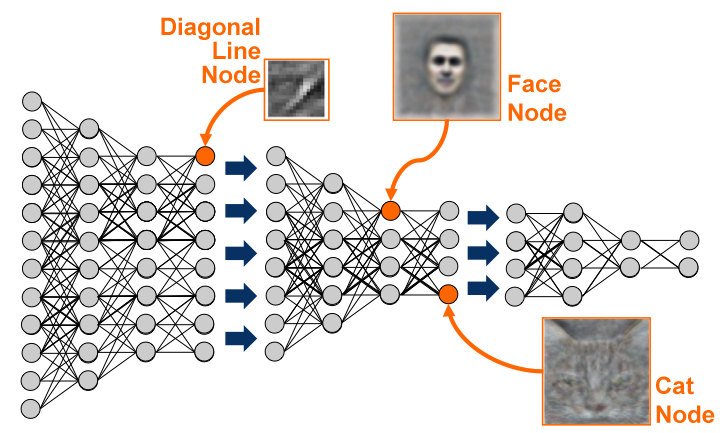}
    \caption{Deep belief network layers learning complex feature hierarchies \cite{cat_visualization}}
    \label{fig:NN_example}
\end{figure}

For example, in a computer vision application, one of the first RBM layers may learn simple features such as to recognize edges.
Following RBM layers may learn a group of edges, such as shapes, whereas top layers may learn to recognize objects, such as faces and so forth. This is visualized in Figure~\ref{fig:NN_example}.

In the second stage, discriminative fine-tuning using backpropagation is performed on the entire DBN to tweak the weights. Because of the pre-training, the weights have a good initialization, which allows backpropagation to quickly optimize the weights, which is explained in detail in \cite{importance_momentum}, \cite{difficulty_training} and \cite{pretraining_help}.

Chapter~\ref{chapter:recommendations} provides practical recommendations on how to configure DBNs, such as the number of hidden layers, the number of hidden units in each layer, how much fine-tuning to perform, etc.

\subsection{Stacked autoencoders}
\label{chapter:deepnn:stacked}
Building a DBN from autoencoders is also called a stacked autoencoder.
The learning process is related to constructing a DBN from RBMs, in particular the layer-wise pre-training, as explained in \cite{stacked_denoising}.
First, an autoencoder is trained on the input. The trained hidden layers serves then as the first hidden layer of the stacked autoencoder. Second, the features learned by the hidden layer are used as input and output to train another autoencoder. The learned hidden layer of the second autoencoder is then used as the second hidden layer of the stacked autoencoder.
This process can be continued for multiple autoencoders, similarly to training a DBN composed of RBMs.
Similarly, each hidden layer learns more complex features.
Last, fine-tuning of the weights using backpropagation is performed on the stacked autoencoder.

\section{Further approaches}
This section briefly covers alternative approaches that are used to train deep neural networks.

\subsection{Discriminative pre-training}
Starting with a single hidden layer, a neural network is trained discriminatively. Subsequently, a second hidden layer is inserted between the first hidden layer and the output layer. Next, the whole network is trained discriminately again. This process can be repeated for more hidden layers. Finally, discriminative fine-tuning is performed in order to tweak the weights. This method is called discriminative fine-tuning and described by Hinton in \cite{hinton_speech}.

\subsection{Hessian-free optimization and sparse initialization}
Based on Newton's method for finding approximations to the roots of a real-valued function, a function $l$ can be minimized by performing Newton's method on its first derivative.
This method can be generalized to several dimensions:
\begin{align}
\theta := \theta - H^{-1} \nabla_\theta l(\theta)
\end{align}

The gradient $\nabla_\theta l(\theta)$ is the vector of partial derivatives of $l(\theta)$ with respect
to the $\theta_i$'s.
The Hessian $H$ is an $n$-by-$n$ matrix of partial derivatives:
\begin{align*}
H_{ij} = \frac{\partial^2 l}{\partial \theta_i \partial \theta_j}
\end{align*}

Because of calculating and inverting the Hessian, each iteration of Newton's method is more expensive than one iteration of gradient descent. Instead, this method converges usually much faster than gradient descent. So-called quasi-Newton methods approximate the Hessian instead of calculating it explicitly.

\cite{hessian_free} presents a so-called Hessian-free approach, which allows to calculate a helper matrix accurately using finite differences and does not require pre-training.
It is combined sparse initialization of weights, i.e. by setting most incoming connection weights to each unit to zero.

\subsection{Reducing internal covariance shift}
As described in \cite{covariance_shift}, during training, a change of the parameters of the previous layers causes the distribution of each layer's input to change. This so-called internal covariance shift slows down training and may result in a neural network that overfits.
Internal covariance shift can be compensated by normalizing the input of every layer. As a consequence, training can be significantly accelerated. The resulting neural network is also less likely to overfit.
This approach is dramatically different to regularization covered in Chapter~\ref{chapter:nnreg}, as it addresses the cause of overfitting, rather than trying to improve a model that overfits.

\chapter{Practical recommendations}
\label{chapter:recommendations}
The concepts described in Chapter~\ref{chapter:deepnn} allow to build powerful deep neural networks in theory.
Yet, there are many open questions on how to tweak them in order to achieve cutting-edge results in applications.
This chapter provides an overview about selected practical recommendations and improvements resulting from them, collected from a variety of publications.
It also includes selected examples of improved classification rates for concrete examples.
Furthermore, this chapter contains elaborations and critical comments by the author of this report on the suggested practical recommendation where necessary.

\section{Activation functions}
Typical neural networks use linear or Sigmoidal activation functions in their output layer for regression or classification, respectively.
For deep neural networks, other activation functions have been proposed, which are covered in this section.
Softmax is a generalization of the Sigmoid activation function, based on the following relation, covered in \cite{hinton_guide}:
\begin{align}
p = \sigma(x) = \frac{1}{1 + e^{-x}} = \frac{e^x(1)}{e^x(1 + e^{-x})} = \frac{e^x}{e^x + e^0}
\end{align}
This can be generalized to $K$ output units:
\begin{align}
p_j = \frac{e^{x_j}}{\sum_{i = 1}^K e^{x_i}}
\end{align}

It is used in the output layer. As described by Norvig in \cite{norvig}, the output of a unit becomes almost deterministic for a unit that is much stronger than the others. Another benefit of softmax is that it is always differentiable for a weight.
As recommended in \cite{learning_deep} and \cite{ng_tutorial}, softmax is ideally combined with the cross-entropy cost function, which is covered in Chapter~\ref{chapter:nntraining}.

In order to model the behavior or neurons more realistically, the rectified linear unit activation function has been proposed in \cite{rectified}:
\begin{align}
f(x) = \max(0, x)
\end{align}
This non-linearity allows to only activate a unit if its output is positive. There is also a smooth approximation of this activation, which is differentiable:
\begin{align}
f(x) = \log(1 + e^x)
\end{align}

Another proposed activation function is a maxout network, which groups the values of multiple units and select the strongest activation, covered in \cite{deep_maxout} and \cite{deng}.

Significant improvements of classification rates using rectified linear units and maxout networks have been reported in \cite{deep_maxout} and \cite{generalized_maxout} for speech recognition and computer vision tasks. In particular, combinations of maxout, convolutional neural networks and dropout, see Chapters~\ref{chapter:nn::other} and \ref{chapter:deepnn:dropout}, respectively, have been reported.
For example, \cite{dropout_simple} compares the performance of different neural networks on the MNIST dataset.
It starts with a standard neural network and a SVM using the radial basis function kernel having error rates of 1.60\% and 1.40\%, respectively.
It then presents various dropout neural networks using rectified linear units with errors around 1.00\% and a dropout neural network using maxout units with error of 0.94\%.
Deep belief networks and so-called deep Boltzmann machines using fine-tuned dropout achieved error rates of 0.92\% and 0.79\%, respectively.

\section{Architectures}
As discussed in Chapter~\ref{chapter:nn:feed}, there are many different (and opposing) guidelines on the architecture of neural networks. The same considerations apply to deep neural networks. In general, the more units and the more layers, the more complex hypotheses can be learned by a network. Considering each hidden layer as a feature detector, the more layers, the  more complex feature detectors can be learned.
As a consequence, a very deep network tends to overfit and therefore requires strong regularization or more data.

Finding a good architecture for a specific task is a model selection problem and is linked to specific training choices made, which are covered in Chapter~\ref{chapter:recommendations::training}.
Different observations and results on architecture evaluations have been reported in the literature, for example in
\cite{hinton_shapes} and \cite{difficulty_training} on MNIST.

For layer-wise pre-training of RBMs, Hinton provides a vague recipe for approximating the number of hidden units of a RBM in \cite{hinton_guide}. No further details on the justification of this recipe are provided.
In order to reduce overfitting, the number of bits to describe a data vector must be approximated first, for example using entropy. Second, this number should be multiplied with the number of training cases. The number of units should then be an order of magnitude smaller of that product.
Hinton adds that for very sparse models, more units may be picked, which makes intuitively sense, as sparsity itself is a form of regularization, as discussed in Chapter~\ref{chapter:autoenc:den}.
Furthermore, he notes that large datasets are prone to high redundancy, for which he recommends less units in order to compensate the risk of overfitting.

\section{Training of RBMs and autoencoders}
\label{chapter:recommendations::training}
This section summarizes recommendations concerning the training of RBMs and autoencoders.
Hinton and Ng provide details on training RBMs and autoencoders in \cite{hinton_guide} and \cite{ng_tutorial}, respectively.

\subsection{Variations of gradient descent in RBMs}
Hinton proposes to use small initial weights in RBMs. Similar to the difficulty of training regular feed-forward networks discussed in Chapter~\ref{chapter:nntraining}, larger initial weights may speed up learning, but would result in a model prone to overfitting.
Concretely, he proposes initial weights drawn from a zero-mean Gaussian distribution with a small standard deviation of 0.01.
For the bias terms, he proposes to set the hidden biases to 0 and the visible biases to $\log(p_i/(1-p_i))$ with $p_i$ being the fraction of training examples in which unit $i$ is activated.

\subsubsection{Mini-batch gradient descent}
Stochastic gradient descent defined in Algorithm~\ref{alg:stochasticgradientdescent} is the choice for training neural networks.
Hinton recommends to use mini-batch gradient descent, which is defined in Algorithm~\ref{alg:minibatchgradientdescent}.

\begin{algorithm}
\caption{Mini-batch gradient descent: training size $m$, learning rate $\alpha$}
\label{alg:minibatchgradientdescent}
\begin{algorithmic}
\State $b\gets $ batch size
\Repeat
\For{$i=1$ to $m$, $step=b$}
\State $\theta_j  \gets \theta_j - \alpha \frac{\partial}{\partial \theta_j}J(\theta, (x^{(i)}, y^{(i)}), ..., (x^{(i+step)}, y^{(i+step)}))$ (simultaneously for all $j$)
\EndFor
\Until{convergence}
\end{algorithmic}
\end{algorithm}

Comparing (batch) gradient descent defined in Algorithm~\ref{alg:gradientdescent} to stochastic gradient descent, it can be concluded that mini-batch gradient descent is a combination of both algorithms.
This algorithm computes the parameters for batches of $b$ training examples.
Mini-batch gradient allows like stochastic gradient descent to move quickly towards a minimum and the possibility to escape from local minima.
In addition it can be vectorized. Vectorization represents  $k$ training examples in a matrix:
\begin{align}
X = \begin{pmatrix}
\mbox{------} (x^{(1)})^T \mbox{------} \\
\mbox{------} (x^{(2)})^T \mbox{------} \\
\vdots \\
\mbox{------} (x^{(k)})^T \mbox{------}
\end{pmatrix}
\end{align}

Gradient descent can then be computed in a sequence of matrix operations.
Efficient algorithms such as the Strassen algorithm defined in \cite{strassen} compute matrix multiplications in $O(n^{log_2 7}) \approx O(n^{2.807})$ instead of the naive multiplication which is $O(n^3)$.
The idea behind the Strassen algorithm is to define a matrix multiplication in terms of recursive block matrix operations. Exploiting redundancy among the block matrices, it results in less multiplications.
The Strassen algorithm comes with runtime performance gains, both because of less multiplications and is also able to be parallelized, making it an ideal candidate for executing in a GPU.

The batch size of the mini-batches must be well balanced. If it is too large, the algorithm's behavior is close to gradient descent and if it is too small, no advantage of vectorization can be taken. Hinton recommends to set the batch size equal the number of classes and to put a training example of each class in each every batch.
If this is not possible, he recommends to randomly shuffle the training set first and then to use a batch size of about 10.
As efficient matrix multiplications come with a large constant, Ng notes in \cite{ng_mlc} that many libraries do not apply it to small matrices, as the naive multiplication is often faster in such cases.
Elaborating on this, it can be concluded that a larger batch size than 10 is probably more useful for runtime performance gains, ideally with a number of training examples close to the number of features and both being slightly less or equal to a power of 2, as the Strassen algorithm expands matrix dimensions to a power of 2.

\subsubsection{Learning rate}
For any gradient-based learning algorithm, setting the learning rate $\alpha$ is crucial. A too small learning rate will make a learning algorithm converge too slowly. In contrast, a too large learning rate will make the learning algorithm overshoot and possibly lead to significant overfitting.
Hinton recommends to generate histograms of weight updates and weight values. The learning rate should then be adjusted to scale the weight updates to about 0.1\% of the weight values, but provides no further justification for that choice.

\subsubsection{Momentum}
Furthermore, Hinton proposes to combine the weight updates with momentum, which allows to speed up learning through long and slowly decreasing "valleys" in a cost function, but increasing the velocity of the weight updates.
In the simplest method, a learning rate could be simply multiplied by $1/(1-\mu)$ in order to speed learning up.
Alternatively, the weight update is then computed, where $\alpha$ is the learning rate:
\begin{align}
\Delta \theta_i(t) = \mu \Delta\theta_i(t-1) - \alpha \frac{\partial J(\Theta)}{\partial \theta_i}(t)
\end{align}
In order to compute the weight update, the previous weight update is taken into account.
The weight update is then accelerated by a factor of $1/(1-\mu)$ if the gradient has not changed.
Hinton notes that this update rule applying temporal smoothing is more reliable than the basic method discussed initially.
He proposes to start with a low momentum value $\mu$ of 0.5 giving an update factor of $1/(1-0.5) = 2$ for parameter updates, as random initial weights may cause large gradients.
A too large momentum value could cause the algorithm to overshoot by oscillating, which is less likely giving a low momentum value.
More details on weight initialization and momentum are provided in \cite{importance_momentum}.

Hinton adds that $L_1$ and $L_2$ regularization can be used as well for the training of RBMs.

\subsection{Autoencoders}
\begin{figure}[h!]
    \centering
    \includegraphics[width=0.6\textwidth]{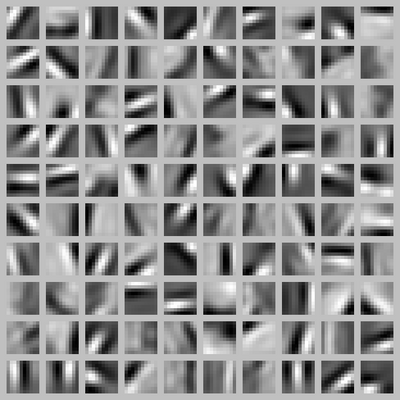}
    \caption{Features learned by 100 hidden units of the same layer \cite{ng_tutorial}}
    \label{fig:features_visual}
\end{figure}

For autoencoders, there have been less recommendations available in the literature. Most prominently, Ng's tutorial in \cite{ng_tutorial} provides advice on training autoencoders.
In particular, that tutorial provides advice on data preprocessing, which helps to improve the performance of deep learning algorithms as stated in the tutorial. It covers mean-normalization, feature scaling and whitening, a decorrelation procedure.
As explained by Ng, whitening is particularly helpful to process image data, which contains large redundancies, and preprocessing led to significantly better results.
Preprocessing then also helps to visualize the features learned by the units, as visualized in Figure~\ref{fig:features_visual}.
Concretely, it shows the input image that maximally activates each of the 100 hidden units.

Furthermore, \cite{difficulty_training} provides a study on the number of hidden layers of stacked autoencoders for the MNIST dataset, with 4 layers being the best tradeoff of over- and underfitting.

\subsection{Fine-tuning of DBNs}
The effect of fine-tuning has been extensively evaluated in \cite{hinton_shapes} on MNIST by Hinton.
He starts with a pre-trained DBN. Pre-trained DBNs usually already have a good classification rate, as their feature detectors are learned from the training data. Hinton highlights that only a small learning rate should be picked in order to not change the network weights too much in the fine-tuning stage.
He then applies backpropagation fine-tuning with early stopping to reduce classification errors to about 1.10\%.

As investigated in \cite{exploring_strategies}, fine-tuning may result in strong overfitting. Whether doing it at all should therefore be subject to model selection, not just the number of epochs and when to stop.

\chapter{Application to computer vision problems}
This chapter applies some of the methods of the previous chapters to real learning problems. Different methods are compared on two databases and their results are presented.

\section{Available databases}
\label{exp::databases}
This section briefly covers the two databases that are of interest for the experiments of this report.

\subsection{MNIST}
\begin{figure}[h!]
    \centering
    \includegraphics[width=0.6\textwidth]{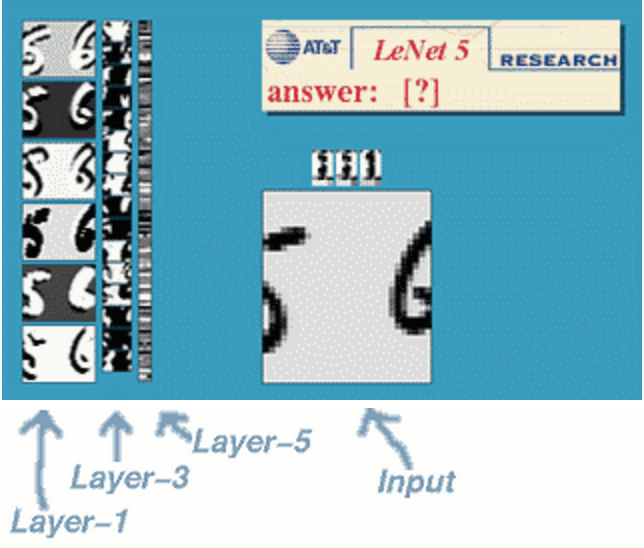}
    \caption{Hand-written digit recognition learned by a convolutional neural network \cite{MNIST_LeNet}}
    \label{fig:LeNet}
\end{figure}

The Mixed National Institute of Standards and Technology (MNIST) database \cite{MNIST} is a collection of handwritten digits with different levels of noise and distortions. It was initially created by LeCun for his research on convolutional neural networks in the 1980s shown in Figure~\ref{fig:LeNet}.

It is used in many papers on deep learning to which there have been numerous references. For example, \cite{dropout_simple} and \cite{hinton_science} report error rates of 1.05\% and 1.2\%, respectively. It is important to bear in mind that these error rates were only achieved applying many different tweaks and optimizations, which are partially covered in Chapter~\ref{chapter:recommendations}.

In the following experiments, the MNIST dataset that comes with the MATLAB Deep Learning Toolbox, introduced in Chapter~\ref{exp::toolbox}, is used. It has 60000 training examples and 10000 test examples. Each example contains $28\times 28$ pixel gray-scale values, of which most are set to zero.

\subsection{Kaggle facial emotion data}
\label{exp::databases::kaggle}
Kaggle is a platform that hosts data mining competitions. Researchers and data scientists compete with their models trained on a given training set in order to achieve the best prediction on a test set.
In 2013, a challenge named "Emotion and identity detection from face images" \cite{kaggle_competition} was hosted.

This challenge was won by a convolutional neural network presented in \cite{tang}, which achieved an error rate of 52.977\%. In total, there were 72 submissions, with 84.733\% being the highest error rate and a median error rate of 63.255\%.

\begin{figure}[h!]
    \centering
    \includegraphics[width=0.6\textwidth]{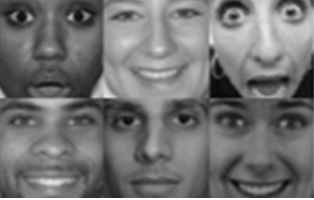}
    \caption{Sample data of the Kaggle competition \cite{kaggle_competition}}
    \label{fig:KaggelFaces}
\end{figure}

The original training set contains 4178 training and 1312 test examples for seven possible emotions. Each example contains $48\times 48$ pixel gray values, of which most are set to non-zero. Since the test labels are not available, the original training set is split up into 3300 training and 800 test examples. Some examples are visualized in Figure~\ref{fig:KaggelFaces}.

\section{Available libraries}
This section briefly covers the two deep learning libraries that are of interest for the experiments of this report.

\subsection{Hinton library}
Based on his paper \cite{hinton_science}, Hinton provides a MATLAB implementation of a stacked autoencoder in \cite{hinton_code}. This code is highly optimized and lacks generalization to other problems. Since this report covers different databases and different deep training methods, it is not further considered.

\subsection{Deep Learning Toolbox}
\label{exp::toolbox}
Palm provides a generic and flexible toolbox for deep learning in \cite{deep_toolbox} based on \cite{palm_publication}. It includes implementations of deep belief networks composed of RBMs, stacked (denoising) autoencoders, convolutional neural networks, convolutional autoencoders and regular neural networks.
Furthermore, it comes with ready to use code examples for MNIST. These examples make the library easy to use and demonstrate various possible configurations.

\section{Experiments}
For the following experiments on the two databases presented in Chapter~\ref{exp::databases}, Palm's Deep Learning Toolbox introduced in Chapter~\ref{exp::toolbox} is used. This section only covers very initial experiments to get a taste of the key training methods covered in this report.

\subsection{Classification of MNIST}
\label{chapter::classification_MNIST}
In this section, deep belief networks composed of RBMs (DBN) are compared to stacked denoising autoencoders (SAE) for the MNIST dataset, covered in Chapters~\ref{chapter:deepnn}.

The MNIST input is normalized from integer values in $[0, 255]$ to real values in $[0, 1]$ and used throughout the following experiments.
For both training methods, a number of parameters are optimized independently. Table~\ref{tab:model_MNIST} contains the different values of parameters that are tested on both training methods throughout pre-training and fine-tuning.

\begin{table}[h!]
\centering
	\begin{tabular}{c || c || c}
		Parameter & Default value & Tested values \\
		\hline
		\hline
		Learning rate & 1.0 & 0.25, 0.5, 0.75, 1.0, 1.25, 1.5, \\
		 & &1.75, 2.0 \\
		Momentum & 0 & 0.01, 0.02, 0.05, 0.1, 0.15, 0.2, \\
		 & & 0.25, 0.5 \\
		$L_2$ regularization & 0 & 1e-7, 5e-7, 1e-6, 5e-6, 1e-5, 5e-5, \\
		 & &1e-4, 5e-4 \\
		Output unit type & Sigmoid & Sigmod, softmax \\
		Batch size & 100 & 25, 50, 100, 150, 200, 400 \\
		Hidden Layers & [100, 100] & [50], [100], [200], [400], [50, 50],  \\
		 & & [100, 100], [200, 200], [400, 400], \\
		 & & [50, 50, 50], [100, 100, 100], \\ 
		 & & [200, 200, 200] \\
		Dropout & 0 & 0, 0.125, 0.25, 0.5 \\
	\end{tabular}
\caption{Model selection values for MNIST}
\label{tab:model_MNIST}
\end{table}

\begin{figure}[h!]
    \centering
    \includegraphics[width=\textwidth]{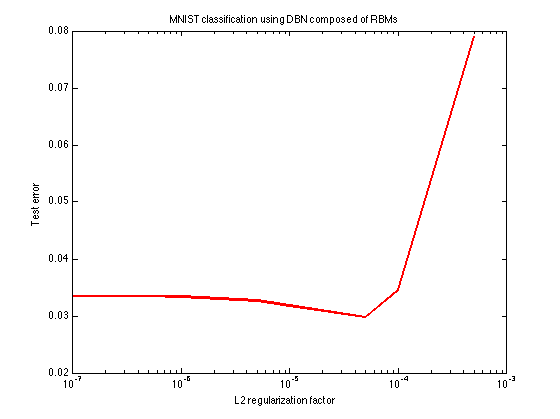}
    \caption{Test error for different $L_2$ regularization values for training of DBN}
    \label{fig:MNIST_RBM_L2}
\end{figure}

The parameters are optimized independently for computational performance reasons. During the optimization, the other parameters are set to the default values in Table~\ref{tab:model_MNIST}.
Figure~\ref{fig:MNIST_RBM_L2} visualizes the test error for different $L_2$ regularization values, with an optimal value for 5e-5 of 0.0298.

The same procedure is followed for the other parameters with 10 epochs for both pre-training and fine-tuning.
Table~\ref{tab:model_RBM_SAE_MNIST} includes the optimal values and respective test errors for both DBN and SAE.
Overall, the test errors are below 4\% for the selected parameter values. Both training methods achieve the best error rate improvement for a change of the number of hidden units from one layer of 100 units to two layers of 400 units each.

\begin{table}[h!]
\centering
	\begin{tabular}{c || c c || c c}
		Parameter & DBN & Test error & SAE & Test error \\
		\hline
		\hline
		Learning rate & 0.5 & 0.0323 & 0.75 & 0.0383 \\
		Momentum & 0.02 & 0.0331 & 0.5 & 0.039 \\
		$L_2$ regularization & 5e-5 & 0.0298 & 5e-5 & 0.0345 \\
		Output unit type & softmax & 0.0278 & softmax & 0.0255 \\
		Batch size &  50 & 0.0314 & 25 & 0.0347 \\
		Hidden Layers & [400, 400] & \textbf{0.0267} & [400, 400] & \textbf{0.017} \\
		Dropout & 0 & 0.0335 & 0 & 0.039 \\
	\end{tabular}
\caption{Model selection for DBN and SAE on MNIST, lowest error rates in bold}
\label{tab:model_RBM_SAE_MNIST}
\end{table}

Finally, the optimal values are put together for both training methods in order to train two optimized classifiers.
In addition, a regular non-denoising stacked autoencoder is trained on these parameter values.
Table~\ref{tab:RBM_SAE_MNIST_res} contains the respective results.

\begin{table}[h!]
\centering
	\begin{tabular}{c || c}
		Neural network & Test error \\
		\hline
		\hline
		DBN composed of RBMs & 0.0244 \\
		Stacked denoising autoencoder & \textbf{0.0194} \\
		Stacked autoencoder & 0.0254
	\end{tabular}
\caption{Error rates for optimized DBN and SAE on MNIST, lowest error rate in bold}
\label{tab:RBM_SAE_MNIST_res}
\end{table}

The stacked denoising autoencoder achieves with a test error of 1.94\% clearly the best result of the three classifiers.
The DBN composed of RBMs performs with an error rate of 2.44\% in comparison to the stacked autoencoder with an error rate of 2.54\%.

Compared to the error rates reported in \cite{dropout_simple} and \cite{hinton_science} of 1.05\% and 1.2\%, the MATLAB Deep Learning Toolbox gets quite close to these best values reported with simple parameter optimization and without any hacks in the implementation limited to the concrete MNIST problem.

For an exhaustive optimization, the error rates are likely to go down further. The single optimization of the architecture in Table~\ref{tab:model_RBM_SAE_MNIST} achieved an error of 1.7\% for the stacked denoising autoencoder in comparison to the aggregation of optimized values in Table~\ref{tab:RBM_SAE_MNIST_res}.
Furthermore, significantly increasing the number of epochs is likely to further reduce the error rates.

\subsection{Classification of Kaggle facial emotion data}
In this section, deep belief networks composed of RBMs (DBN) are compared to stacked denoising autoencoders (SAE) for the Kaggle facial emotion dataset, covered in Chapters~\ref{chapter:deepnn}.

Each data point has $48\times 48 = 2304$ pixels. As there are only 3300 training examples, initial experiments returned impractical error rates of 90\%.
One of the reasons is the low ratio of the number of training examples and the number of features.
Therefore, the image size is reduced to $24\times 24 = 576$ pixels using a bilinear interpolation. Using the nearest 2-by-2 neighborhood, the result pixel value is a weighted average of pixels in the neighborhood.
Subsequently, the input is normalized from integer values in $[0, 255]$ to real values in $[0, 1]$ and used throughout the following experiments.

Similar to the model selection for MNIST in Chapter~\ref{chapter::classification_MNIST}, for both training methods, a number of parameters are optimized independently. Table~\ref{tab:model_Kaggle} contains the different values of parameters that are tested on both training methods throughout pre-training and fine-tuning.

\begin{table}[h!]
\centering
	\begin{tabular}{c || c || c}
		Parameter & Default value & Tested values \\
		\hline
		\hline
		Learning rate & 1.0 & 0.05, 0.1, 0.15, 0.25, 0.5, 0.75, 1.0, \\ 
		 & & 1.25, 1.5 \\
		Momentum & 0 & 0.01, 0.02, 0.05, 0.1, 0.15, 0.2, \\
		 & & 0.25, 0.5 \\
		$L_2$ regularization & 0 & 1e-7, 5e-7, 1e-6, 5e-6, 1e-5, 5e-5, \\
		 & & 1e-4, 5e-4 \\
		Output unit type & Sigmoid & Sigmod, softmax \\
		Batch size & 100 & 25, 50, 100, 150, 275 \\
		Hidden Layers & [100, 100] & [50], [100], [200], [400], [50, 50], \\
		 & & [100, 100], [200, 200], [400, 400], \\
		 & & [50, 50, 50],  [100, 100, 100], \\
		 & & [200, 200, 200] \\
		Dropout & 0 & 0, 0.125, 0.25, 0.5 \\
	\end{tabular}
\caption{Model selection values for Kaggle data}
\label{tab:model_Kaggle}
\end{table}

\begin{figure}[h!]
    \centering
    \includegraphics[width=\textwidth]{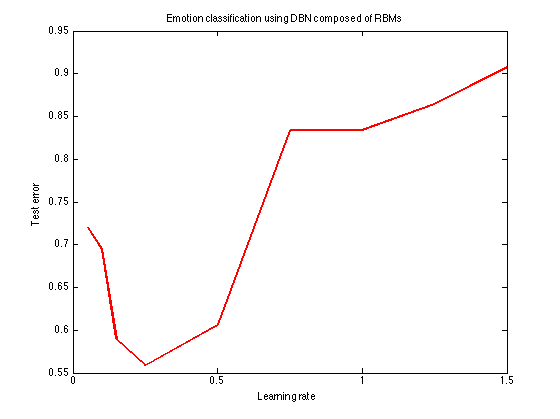}
    \caption{Test error for different learning rates values for training of DBN}
    \label{fig:MNIST_RBM_alpha}
\end{figure}

The parameters are also optimized independently for computational performance reasons. During the optimization, the other parameters are set to the default values in Table~\ref{tab:model_Kaggle}.
Figure~\ref{fig:MNIST_RBM_alpha} visualizes the test error for different learning rates, with an optimal value for 0.1 of 0.5413.

The same procedure is followed for the other parameters with 10 epochs for both pre-training and fine-tuning.
Table~\ref{tab:model_RBM_SAE_Kaggle} includes the optimal values and respective test errors for both DBN and SAE.
Overall, the test errors are high for the selected parameter values. As discussed in Chapter~\ref{exp::databases::kaggle}, the best contribution has an error rate of 52.977\%.
Also, in many instances of the model selection, the test error plateaus on 72.25\%. This is most likely caused by the noise and redundancy in the data set and the overall low amount of training examples.
Possible solutions are discussed at the end of this experiment.

\begin{table}[h!]
\centering
	\begin{tabular}{c || c c || c c}
		Parameter & DBN & Test error & SAE & Test error \\
		\hline
		\hline
		Learning rate & 0.25 & \textbf{0.5587} & 0.1 & 0.5413 \\
		Momentum & 0.01 & 0.7225 & 0.5 & 0.7225 \\
		$L_2$ regularization & 5e-5 & 0.7225 & 1e-4 & 0.7225 \\
		Output unit type & softmax & 0.7225 & softmax & 0.7225 \\
		Batch size &  50 & 0.6987 & 50 & 0.5913 \\
		Hidden Layers & [50, 50] & 0.7225 & [200] & \textbf{0.5850} \\
		Dropout & 0.125 & 0.7225 & 0.5 & 0.7225 \\
	\end{tabular}
\caption{Model selection for DBN and SAE on Kaggle data, lowest error rates in bold}
\label{tab:model_RBM_SAE_Kaggle}
\end{table}

Finally, the optimal values are put together for both training methods in order to train two optimized classifiers.
In addition, a regular non-denoising stacked autoencoder is trained on these parameter values.
Table~\ref{tab:RBM_SAE_Kaggle_res} contains the respective results.

\begin{table}[h!]
\centering
	\begin{tabular}{c || c}
		Neural network & Test error \\
		\hline
		\hline
		DBN composed of RBMs & 0.7225 \\
		Stacked denoising autoencoder & 0.7225 \\
		Stacked autoencoder & \textbf{0.3975}
	\end{tabular}
\caption{Error rates for optimized DBN and SAE on Kaggle data, lowest error rate in bold}
\label{tab:RBM_SAE_Kaggle_res}
\end{table}

Both, the DBN composed of RBMs and the stacked denoising autoencoder plateau at an error rate of 72.25\%.
Interestingly, using the same parameter values of the stacked denoising autoencoder for a regular stacked autoencoder, the test error goes down to 39.75\%.
This is clearly a major advancement to the best Kaggle contribution with an error of 52.977\%.
However, it must be noted that the test set is not identical, since the test labels are not available, which required the original training set to be split into training and test data. Bearing this in mind, the learning problem in this experiment may even be harder since less training examples are available.

More advanced pre-processing methods are the Principal Component Analysis (PCA) or whitening discussed and recommended for gray-scale images in \cite{ng_tutorial}.
Using PCA, only a relevant fraction of the features can be used for the learning in order to reduce the plateau of 72.25\% and to speed up learning.
Gray-scale images are highly-redundant and using whitening, the features can be decorrelated in order to further improve error rates.
Furthermore, limitations in the implementation of the toolbox may also contribute to the high plateau of error rates. For example, linear units in the visible layer of the RBMs could also handle the input better and may improve error rates.
Also, as for the MINIST dataset, significantly increasing the number of epochs is likely to further reduce the error rates.

\chapter{Conclusions and prospects}
Neural networks have a long history in machine learning that came with repeating rise and enthusiasm followed by loss popularity. Neural networks are known to have a high expressional power in order to model complex non-linear hypotheses.
In practice, training deep neural networks using the backpropagation algorithm is difficult.
Their large number of weight parameters result in highly non-convex cost functions that are difficult to minimize. As a consequence, training often converges to local minima, resulting in overfitting and lower generalization errors.

Over the last ten years, a number of new training methods have been developed in order to pre-train neural networks resulting in a good initialization of their weights. These weights can then be fine-tuned using backpropagation.
The most popular pre-training methods in order to construct deep neural networks are Restricted Boltzmann Machines (RBMs) and Autoencoders. Aside from these concepts, there are other methods and approaches such as discriminative pre-training and dropout regularization.
Furthermore, there are many practical recommendations to tweak these methods.

As shown in the experiments, deep neural network training methods are powerful and achieve very high classification rates in different computer vision problems. Nonetheless, deep learning is not a concept that can be used out of the box, as proper pre-processing of the data and extensive model selection are necessary.

It would be interesting to apply more powerful pre-processing methods to the data, such as PCA or whitening. Furthermore, convolutional neural networks have been reported in the literature to perform very well on computer vision problems, which would be interesting to study, too.
In order to speed up learning, use of GPUs has been successfully used in the literature. These methods also look promising to be applied to the problems covered in this report.

\end{document}